\documentclass[10pt,twocolumn,letterpaper]{article}

\usepackage{iccv}
\usepackage{times}
\usepackage{epsfig}
\usepackage{graphicx}
\usepackage{amsmath}
\usepackage{amssymb}

\usepackage[english]{babel}
\usepackage[T1]{fontenc}
\usepackage[utf8]{inputenc}
\usepackage{booktabs} 
\newcommand{\norm}[1]{\left\lVert#1\right\rVert}
\newcommand\blfootnote[1]{%
  \begingroup
  \renewcommand\thefootnote{}\footnote{#1}%
  \addtocounter{footnote}{-1}%
  \endgroup
}

\usepackage[pagebackref=true,breaklinks=true,letterpaper=true,colorlinks,bookmarks=false]{hyperref}

\iccvfinalcopy 

\def\iccvPaperID{9432} 

\ificcvfinal\fi

\begin{document}

\title{MRI-GAN: A Generalized Approach to Detect DeepFakes using Perceptual Image Assessment}

\author{Pratikkumar Prajapati$^{\dag}$\\
{\tt\small pratik.prajapati@gmail.com}

\and
Dr. Chris Pollett$^{\dag}$\\
{\tt\small chris@pollett.org}
}

\maketitle
\ificcvfinal\thispagestyle{empty}\fi

\begin{abstract}
DeepFakes are synthetic videos generated by swapping a face of an original image with the face of somebody else. In this paper, we describe our work to develop general, deep learning-based models to classify DeepFake content. We propose a novel framework for using Generative Adversarial Network (GAN)-based models, we call MRI-GAN, that utilizes perceptual differences in images to detect synthesized videos. We test our MRI-GAN approach and a plain-frames-based model using the DeepFake Detection Challenge Dataset. Our plain frames-based-model achieves 91\% test accuracy and a model which uses our MRI-GAN framework with Structural Similarity Index Measurement (SSIM) for the perceptual differences achieves 74\% test accuracy. The results of MRI-GAN are preliminary and may be improved further by modifying the choice of loss function, tuning hyper-parameters, or by using a more advanced perceptual similarity metric.
\end{abstract}

\blfootnote{\dag Department of Computer Science, San José State University, San José, California, USA}
\section{Introduction}

Fake media based on political leaders, cinema actors, and other celebrities have been used to spread propaganda and to damage reputations. For example, a rather straightforward slow down of a speech of Nancy Pelosi was recently used to imply she was slurring her words, and therefore, may be unfit for office. Wonder Woman reboot scenes with Gal Gadot swapped with Linda Carter have taken the internet by storm. In response, to Facebook's refusal to take down the Nancy Pelosi video, a fake video of Mark Zuckerberg emerged where he claims to own Facebook users. It is not hard to imagine a fake video of a political leader announcing something horrific, leading to a real-world catastrophe. In this research, we explore ways to detect such fake content. Our goal is to quantify the accuracy of DeepFake detection for a variety of natural settings using deep learning-based techniques.

There are many kinds of fake media. The kind this paper focuses on is where the identity of one person is swapped with another. This kind of fake media is called a DeepFake. DeepFakes are generally created using two main kinds of neural networks: Generative Adversarial Networks (GANs) and Autoencoders (AEs)~\cite{Goodfellow-et-al-2016, goodfellow2014generative}. GAN-based models are popular because of their ability to generate realistic looking DeepFakes~\cite{Tolosana_2020, Mirsky_2021}.

We have used the DeepFake Detection Challenge (DFDC) dataset~\cite{dolhansky2020deepfake} in our research to train our proposed models. The DFDC dataset was chosen as it contains DeepFake samples generated with many state-of-the-art methods, contains subjects with rich ethnic diversity, and also the video samples are recorded in a variety of natural settings. The hope is using this dataset for training will result in a trained model that can be used to detect DeepFakes generated by several different methods.

In our research, we developed two fake detection methods and compare them. The first method detects faces on a frame by frame basis, and for each such face detected in a frame, uses a Convolutional Neural Network (CNN) model to determine if a face is fake or not. Based on the fake faces found in a sequence of frames, the video is classified as fake or not using simple aggregation algorithm.

The second proposed method is one particular example of a GAN-based approach that we call MRI-GAN. MRI-GAN extends the Face X-Ray ideas of Li~\etal~\cite{li2019face}. Their approach was to compute, using facial landmarks, a soft mask for the facial boundary likely to have been manipulated. They then use a CNN that takes the input image and this boundary to determine if the image has been manipulated. Our approach is to train a GAN on pairs of images $(I, I_m)$. For training, images $I_m$ are generated by applying, a perceptual dissimilarity function $f$, the component of our system that can be varied, as $I_m = f(I,I_{\mbox{related}})$.  Here $I$ is an original image that was not manipulated and $I_{\mbox{related}}$ is a related image, either the original again or a DeepFake. We then use a second network that takes a sequence of frames generated from our GAN and classifies the video as fake or not. 
Our approach is to take a perceptual similarity function $f$, the component of our system that can be varied, then train a GAN on images $I$, both real and fake, to output $f(I,I_{m})$. Here $I_m$ is an original image, which is just $I$ in the case that the image wasn't manipulated and which is the image $I$ was manipulated from if $I$ is a fake. We then use a second network that takes a sequence of frames generated from our GAN and classifies the video as fake or not. 
For our experiments with this approach, in this paper, we use Structural Similarity Index Measurement (SSIM)~\cite{wangbovik2004} for $f$. Our code is publicly available at \url{https://github.com/pratikpv/mri_gan_deepfake}.

We now discuss the organization of the rest of the paper. In the next section, we survey related work and the relationship of our model to it. This is followed by a discussion of the two methods we use to detect DeepFakes. Next we discuss design of the MRI-GAN neural network, the loss functions we use,  and a specific dataset generation process to train the MRI-GAN. We then explain our evaluation methods and experimental results for MRI-GAN and our other models. Finally, in the last section, we draw some conclusion based on our experiments.

\section{Related work}

DeepFakes are generally created with off-the-shelf models including but not limited to StyleGAN~\cite{karras2019analyzing}, CycleGAN~\cite{zhu2017unpaired}, FaceSwapGAN~\cite{faceswap_GAN} and are sometimes combined with some post-processing of the images and the videos such as color balancing to make the fake video look realistic. Early works on the detection of DeepFakes often used different matrices to report outcomes such as accuracy, Area Under the Curve (AUC), Equal Error Rate (EER), etc. and they used different datasets making it hard to compare results~\cite{Tolosana_2020}. To create a more standardized benchmark, Facebook launched the DeepFake detection challenge (DFDC)~\cite{kaggle_dfdc,fb_dfdc}, and provided a dataset for the competition~\cite{dolhansky2020deepfake}. We have used this dataset in our research. 

Seferbekov~\cite{selimsef, kaggle_dfdc}, the winner of the competition, achieved 82\% test accuracy on the complete DFDC test dataset. We closely followed Seferbekov's solution for our plain-frames-based approach. We use this method as a benchmark to compare against our MRI-GAN based method. However, we use a different kind of data prepossessing than Seferbekov. Also, we used Efficient Net B0~\cite{rw2019timm, tan2020efficientnet} to extract features to accommodate for computation power we had rather than Efficient Net B7, which is more computationally intensive. Since we used a simpler Efficient Net, we used different parameters on the number of fake and real faces detected to say whether a video was fake or not. These parameters were determined by a grid search.

Our MRI-GAN method is related to Li~\etal~\cite{Li_2020}. As we have already mentioned in the introduction, they used a landmark-based masking approach they call taking the X-Ray of the image to detect DeepFakes. This X-Ray can be thought of as computing regions to attend to in an image to see where blending of an original image with a modification occurred. This differentiates it from approaches that directly are trained to look for particular manipulation approaches. Like Li~\etal approach also is not trained for a particular manipulation technique. Unlike Li~\etal~\cite{Li_2020}, our model can generate 3-channel artifacts from the source image. Also, our model does not generate a mask as a convex-hull, but instead an image representing predicted perceptual differences between the images and a real face is computed at the pixel level. We chose the name MRI for the images we output as like a MRI over an X-Ray for medical imaging, our MRI images provide more nuanced details than an X-Ray mask. Li~\etal used the preview version of DFDC dataset for testing, not the full dataset, and reported an AUC of 80.92. We used a final version of DFDC dataset for training and validation. The complete test dataset was not made public, so we used the publicly available test set.

Our MRI-GAN may be considered an extension of the image-to-image translation architecture (known as Pix2Pix) described by Isola~\etal~\cite{pix2pix2017}. We use a similar model architecture for both the generative and discriminating models, but our objective function is quite different. The Pix2Pix GAN is a general-purpose GAN architecture used to translate one image into other. We have extended the idea of Pix2Pix GAN to translate a picture of a person's face into its MRI  using custom optimization goal. In that sense, MRI-GAN may also be seen as a special case of Pix2Pix model optimized for MRI generation to detect DeepFakes. 

\section{Methods} \label{section_methods}

This paper compares our two methods to detect DeepFakes. The overall steps for the first method is as follow: We apply various augmentation and distractions to the training set of the DFDC dataset. Then using a pre-trained Multi-task Cascaded Convolutional Networks (MTCNN)~\cite{Zhang_2016, facenet_pytorch}, we locate faces from frames and extract them. To save some computational cost, we extract faces from every tenth frame. Also, we extract each faces detected in the extracted frames. Each frame can have zero or more number of faces. We use a pre-trained Efficient-Net B0 CNN~\cite{rw2019timm, tan2020efficientnet} to extract CNN-based features for each face. These features are passed to Multi-Layer Perceptron (MLP)-based model to classify whether the face is either DeepFake or real. Then results of each frame are aggregated to predict the genuineness of the entire video. We refer this method as plain-frames-based method.

The second method differs a little bit in process: Before we pass images of faces to Efficient-Net B0 CNN, we pass each face to our proposed trained MRI-GAN to predict the MRI of the input face. Then the predicted MRI of the input image is passed to Efficient-Net B0 CNN and the rest of the process is same as first method. We call this method the MRI-based method to detect DeepFakes. We have trained our MRI-GAN to predict the MRI of an input human face image. Section~\ref{section_mri-gan} describes this process in detail.

The DFDC test and validation set make use of data augmentation and distraction methods. The training set of DFDC has no such methods applied. Augmentation and distraction methods were used so that results would generalize better. For augmentation, various types of noise were added to each frame of the videos. Types of noise implemented were Gaussian, Speckle, Salt-and-Pepper, Pepper, Salt, Poisson, and Localvar. We also applied other augmentation methods such as blur, rotation, horizontal flip, re-scale, brightness, and contrast. For distraction, we implemented static, rolling, and spontaneous methods. Each method includes text and geometric shapes as an object to serve as a distraction. When text is used as a distraction, a random alpha-numeric text of length eight is generated and overlaid on the video. The font size of the text is chosen randomly from the scaling of one to six units. The font thickness is chosen randomly from the scaling of one to three units. The font colors are chosen randomly from red, blue, green, white, and black. In the case of static text, a random text is generated and it is overlaid on the video at a fixed location. The location is chosen randomly and fixed for the entire video. In the rolling text scenario, a random text moves from a fixed direction for the duration of the video. The starting location of the text is chosen randomly at the start of the video and the rolling direction may be any of the following: right-to-left, left-to-right, up-to-down, down-to-up. Our spontaneous text method overlays a random text at random locations at random times in the video. All the various options implemented for text-based distraction apply to shapes-based-distractions as well. Circle and rectangle shapes are chosen for the shapes-based-distractions method. We have applied various combinations of random augmentation and random distractions. So a video might have one of more random augmentations applied as well as one or more random distractions also applied to it.

We have applied the augmentation and distraction methods to portion of the DFDC training set before we used MTCNN to detect faces in the frames. We chose this approach to simulate real-life scenario of typical videos which may be of low-quality, noisy, blurry, with distractions or combination of any such things. Because of this, the MTCNN fails to detect faces in some of the videos. In such cases, we simply drop such videos in our training and quantification. Since we have used same processed data for both of our methods, we can compare their results side-by-side. We do not claim that any of our methods achieve higher accuracy than the winning solution~\cite{selimsef} of DFDC. This is because we have not tested our models with private test dataset of DFDC, which is not available publicly~\cite{dolhansky2020deepfake,fb_dfdc,kaggle_dfdc}. Also, we have simply dropped some of the videos in entire dataset in which MTCNN fails to detect any faces. We use only the visual information from videos to detect DeepFakes, audio channels are discarded.

\section{MRI-GAN} \label{section_mri-gan}

As MRI-GAN is our core contribution in this research, we extensively discuss the details of it in this section. MRI-GAN is designed and trained to generate a blank (black) image if the input face of the person is not synthesized, and an image with artifacts if the input face is synthesized. This generated image, we call the image MRI, is used to detect if the input image is fake or not.

The core philosophy behind MRI-GAN is to use perceptual similarity of images to identify if the given video or image is DeepFake or not. MRI-GAN is a framework that utilities how images are perceived by the humans. To experiment with this idea we used the Structural Similarity Index Measurement (SSIM)~\cite{wangbovik2004} as a metric to train the MRI-GAN. However, our approach could also be used with other perceptual similarity measures such as those surveyed by Zhai and Min~\cite{zhai2020perceptual}.

\subsection{Design}

\begin{figure}[h]
    \centering
    \includegraphics[scale=0.5]{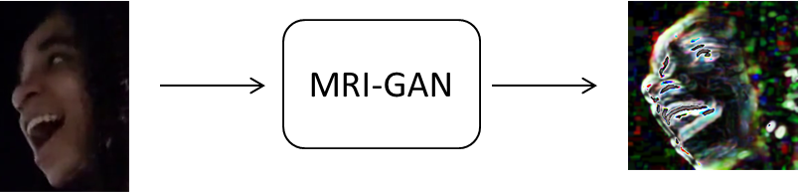}
    \caption{The goal of the MRI-GAN is to generate MRI of face of a person. If the face is DeepFake the MRI will have artifacts otherwise it will be a black image}
    \label{fig:mri_model_arch}
\end{figure}

\begin{figure}[h]
    \centering
    \includegraphics[scale=0.45]{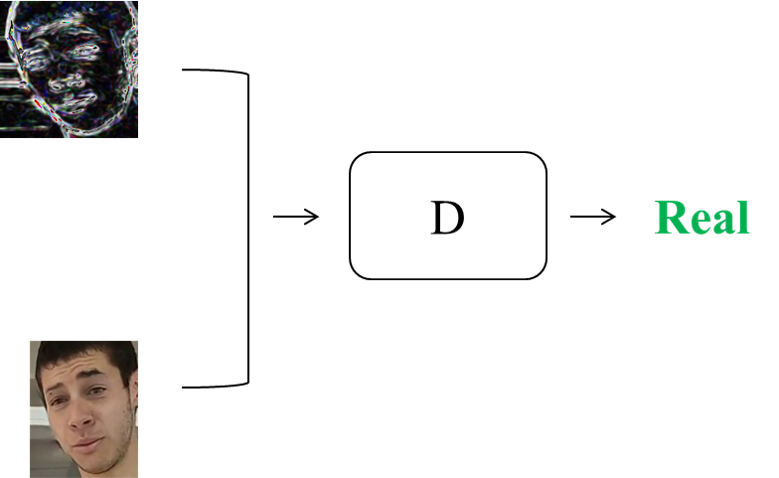}
    \caption{The overall goal of the discriminator model of the MRI-GAN. It outputs zero (or "Real") if the input MRI is of the companion image given as input and one (or "Fake") otherwise.}
    \label{fig:dis_model}
\end{figure}

\begin{figure}[h]
    \centering
    \includegraphics[width=0.47\textwidth]{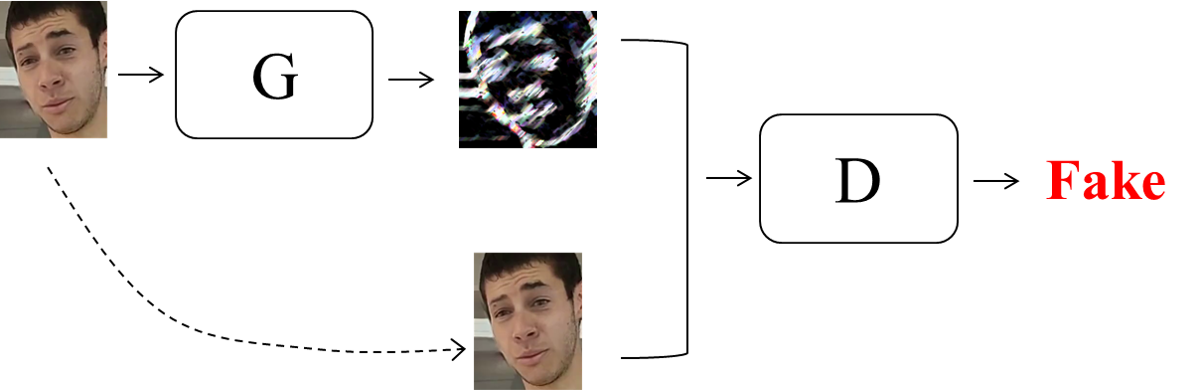}
    \caption{The gverall goal of the generator model of the MRI-GAN.}
    \label{fig:gen_model}
\end{figure}

Our MRI-GAN model takes as input a human face and from it generates and 256 x 256 image we call its MRI. Fig.~\ref{fig:mri_model_arch} illustrates this process.
To train our MRI-GAN model, we have generated a custom dataset, we call the MRI-Deep-Fake dataset (MRI-DF dataset). The detailed process of generating this dataset is described in Section~\ref{section_mri_df_dataset_gen}. In a nutshell this dataset consists of a set of tuples $(I,I_{m})$, where the $I$ is the actual image of a human face and the $I_m$ is its MRI. The image $I$ can be of either real person or a DeepFake image. MRI-GAN closely follows the Pix2Pix architecture described in~\cite{pix2pix2017}, however, our objective function, described in Section.~\ref{objective_function}, is different. Our MRI-GAN consists of a discriminating sub-model and a generating sub-model. The discriminator takes in a set of tuples of form $(I,I_{m})$ and outputs zero (or "Real") to indicate the image $I_m$ is actually the MRI of the image $I$ and one (or "Fake") otherwise.  Fig.~\ref{fig:dis_model} shows the overall goal of the discriminator. The generator model is trained to generate the MRI $I_m$ of the given input image $I$. The discriminator is provided with generated MRI $I_m$ and the actual image $I$ given to the generator, and trained to output zero (or "Fake") to indicate the MRI $I_m$ is generated by the generator model and is not real. Fig.~\ref{fig:gen_model} shows the overall goal of the generator. In Section~\ref{objective_function} we will explain our objective function to back propagate the errors in MRI generation training. The generator and the discriminator is trained in adversarial manner such that goal of the generator is to fool the discriminator by producing $I_m$ which looks perceptually similar to the one sampled from the dataset~\cite{goodfellow2014generative,Goodfellow-et-al-2016}.

\subsection{Perceptual Image Quality Assessment}

Wang~\etal~\cite{wangbovik2004} survey image quality measurement frameworks and suggest that the human visual system (HVS) uses structural information in scenes to assess the perceptual differences between them. They propose SSIM~\cite{wangbovik2004} as a method to quantify the perceived similarity between two images. They suggest SSIM is a better way of measuring the similarity between images as compared to Mean Squared Error (MSE) and Peak signal-to-noise ratio (PSNR). This is because MSE and PSNR use absolute errors while SSIM makes use of the perceived quality of the image. In this paper, we have implemented MRI-GAN and derived our MRI-DF dataset using SSIM as the perceptual similarity function. For this reason, we now briefly explain SSIM, a detailed explanation can be found at Wang \etal~\cite{wangbovik2004}.

SSIM is computed in terms of three components: luminance,  contrast, and structure. Given two images $X$ and $Y$ of the same dimensions, each of the components is calculated for each $(x,y)$ pixel location in these images with respect to a particular color channel and an $N \times N$ neighborhood of pixels of which $(x,y)$ is the center. We imagine $0$-padding the $N \times N$ neighborhood if we are at an image boundary. A global score can be had by then taking the mean across all pixels. For our purposes, though, we are actually interested in making new images using the $(x,y)$ SSIM scores for each pixel location. In this paper, we denote the normalized scalar value of similarity between two images $X$ and $Y$, as $SSIM_{index}(X,Y)$ and pixel-wise similarity as $SSIM_{image}(X,Y)$, the former is a scalar and the latter, an image with pixel values $SSIM_{X,Y}(x,y)$. We now discuss each of the component scores in turn and how they can be used to make a final $SSIM_{X,Y}(x,y)$ formula. When the images $X,Y$ are understood we will write $SSIM(x,y)$ for $SSIM_{X,Y}(x,y)$.

To compute luminance, the mean intensity of pixel values in each coordinate direction center at $(x,y)$ is calculated:
\begin{equation}\label{eq_lum}
\mu_x = \frac{1}{N}\sum_{i=1}^{N}  x_i \quad ,  \quad \mu_y = \frac{1}{N}\sum_{i=1}^{N}  y_i
\end{equation}
The similarity between the luminance of two images $X$ and $Y$, $l(x,y)$, can be estimated as:
\begin{equation} \label{eq_lum_comp}
l(x,y) = \frac{2 \mu_x \mu_y + C_1}{\mu_x^2 + \mu_y^2 + C_1}
\end{equation}
Here $C_1$ is a constant used to avoid instability when denominator the Eq.~\ref{eq_lum_comp} is very close to zero. Wang \etal~\cite{wangbovik2004} suggest using $C_1 = (K_1 L)^2$, where $L$ is the dynamic range of the pixel values (0 to 255) and ${K_1 \ll 1}$.

To compute contrast, the standard deviation of the pixel values in the image patch in each direction are calculated:
$$\label{eq_contr}
\sigma_x = \Big( \frac{1}{N-1}\sum_{i=1}^{N}  (x_i - \mu_x )^2\Big)^\frac{1}{2},$$
$$\sigma_y = \Big( \frac{1}{N-1}\sum_{i=1}^{N}  (y_i - \mu_y )^2\Big)^\frac{1}{2}$$
Like $l(X,Y)$, from this, the contrast comparison function $c(x,y)$ is defined:
\begin{equation} \label{eq_contr_comp}
c(x,y) = \frac{2\sigma_x \sigma_y + C_2}{\sigma_x^2  + \sigma_y^2 + C_2}
\end{equation}
where $C_2 = (K_2 L)^2$ and ${K_2 \ll 1}$.
The structure comparison function $s(x,y)$ is computed in terms of the deviations above and a covariance in the directions: 
\begin{equation} \label{eq_struct_comp}
s(x,y) = \frac{\sigma_{xy} + C_3}{\sigma_x \sigma_y + C_3}
\end{equation}
\begin{equation} \label{eq_sigma_xy}
where, \sigma_{xy} = \frac{1}{N-1}\Big(\sum_{i=1}^N (x_i-\mu_x) (y_i-\mu_y)\Big)
\end{equation}
Finally, $SSIM(x,y)$ is defined by Eq.~\ref{eq_struct_comp1}.
\begin{equation} \label{eq_struct_comp1}
SSIM (x,y) = [l(x,y)]^\alpha \cdot [c(x,y)]^\beta \cdot [s(x,y)]^\gamma
\end{equation}
where, ${\alpha > 0}$,  ${\beta > 0}$, and ${\gamma > 0}$ are used to set relative weights of the three components. Wang \etal~\cite{wangbovik2004} have set ${ \alpha = \beta = \gamma = 1}$ and ${c_3 = \dfrac{c_2}{2}}$ to simplify the $SSIM(x,y)$ equation to:
\begin{equation} \label{eq_SSIM}
SSIM (x,y) = \frac{(2\mu_x\mu_y + C_1)(2\sigma_{xy} + C_2)}{(\mu_x^2  + \mu_y^2 + C_1) (\sigma_x^2  + \sigma_y^2 + C_2)}
\end{equation}
We used the well-vetted scikit-image library~\cite{scikit-image} to compute $SSIM_{index}$ and $SSIM_{image}$ and for our experiments $N=11$.

\subsection{The Definition of MRI of an Image}
For this paper, we compute perceptual structural differences between two images using SSIM via the equation,
\begin{equation}  \label{eq_MRI}
\begin{aligned}
    MRI_{image}(X,Y) = 1 - SSIM_{image}(X,Y)
\end{aligned}
\end{equation}
The image computed by Eq.~\ref{eq_MRI} when $X$ is a real image and $Y$ is an image derived from $X$ is what we call the {\bf ground truth MRI image} of $Y$. Here 
$SSIM_{image}(X,Y)$ is a matrix of dimension $N$ x $N$. When $X = Y$, $MRI_{image}(X,Y)$ will be a matrix of all zeros. I.e., a blank, black image. On the other hand, if $Y$ is a DeepFake image of $X$ then the MRI will represent the perceptual difference between corresponding real and DeepFake image. Like we mentioned previously, the $SSIM_{image}(X,Y)$ outputs a 3-channel, color image, so the $MRI_{image}(X,Y)$ is also 3-channel, color image. Fig.~\ref{fig:MRI_DF_dataset_gen} and Fig.~\ref{fig:MRI_demo_image} show some examples of MRI images.

\subsection{The MRI-DF Dataset} \label{section_mri_df_dataset_gen}

\begin{figure}[h!]
    \centering
    \includegraphics[width=0.47\textwidth]{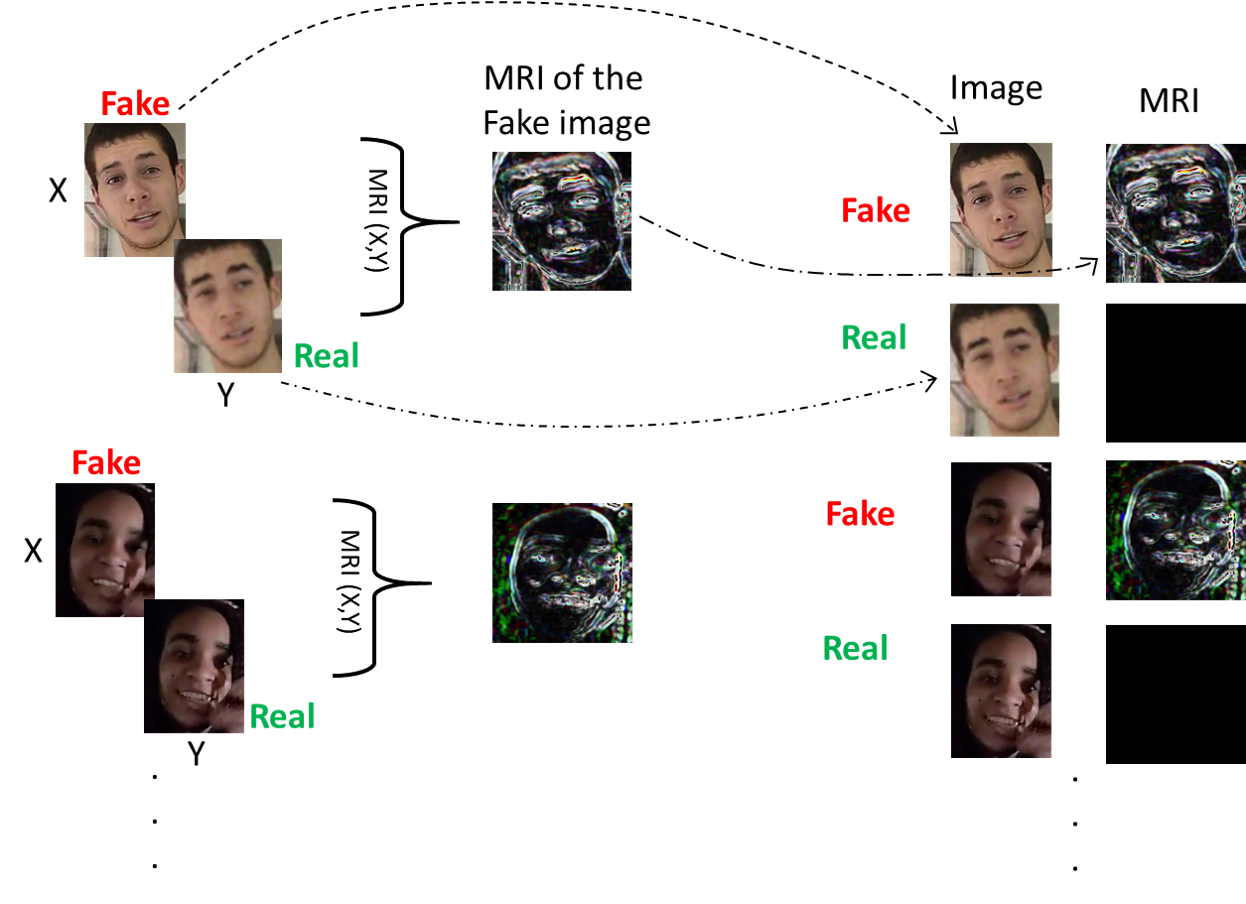}
    \caption{The MRI-DF dataset generation process. The MRI between fake and real face is generated using SSIM and used as the MRI of the fake image. A blank, black image is used as the MRI of the real face. The pairs (image, MRI) are used to train the MRI-GAN}
    \label{fig:MRI_DF_dataset_gen}
\end{figure}

MRI-GAN training data consists of image pairs consisting of an image of a face and its MRI. If the face is not synthesized the MRI is a blank, black image, and if the face is a fake, the MRI will be a structural difference between the fake image and corresponding real image. To make a dataset of such pairs, we randonly took  50\% of the videos in DFDC training set and all of the videos in Celeb-DF-v2 dataset~\cite{Celeb_DF_cvpr20}. Only 50\% of the data from DFDC training set was used so that each dataset was roughly equally represented. As these dataset have more fake content than real content, we used images of real faces from FDF~\cite{10.1007/978-3-030-33720-9_44} and FFHQ~\cite{ffhq_dataset}. The DFDC and Celeb-DF-v2 dataset contain metadata information about the mapping between fake video clips and the real ones. We use this information to generate the MRIs between frames of fake and real video clips. We used blank, black images as MRIs for frames from the real video clips. We call the resulting data from doing the above processing the MRI-DF dataset. Fig.~\ref{fig:MRI_DF_dataset_gen} shows the process pictorially. In addition to the above, we have applied various augmentation and distraction methods as described in Section~\ref{section_methods} on the DFDC data before calculating SSIMs to generate the MRI of the images. We have not applied any augmentation and distraction methods on Celeb-DF-v2, FDF, and FFHQ dataset to derive the MRI-DF dataset.

\subsection{Objective function} \label{objective_function}
Training our MRI-GAN involved optimizing the loss functions used for the discriminator and the generator models independently. For our generator model, our overall loss function was the sum of three components:

\begin{enumerate}
   \item Conditional GAN loss: $\mathcal{L}_{cGAN}(G)$
   \item Pixel-wise L2 loss: $\mathcal{L}_{L_{2}}(G)$
   \item Perceptual loss: $\mathcal{L}_{per}(G)$
\end{enumerate}

$\mathcal{L}_{cGAN}(G)$ is defined by Mirza and Osindero in ~\cite{mirza2014conditional} as:
\begin{equation} \label{cGAN_equation}
\begin{aligned}
   \mathcal{L}_{cGAN}(G,D) = \mathbb{E}_{{x} \sim p_{\text{data}} ({x})}[\log D({x} | {y})] +  \\   
                                \mathbb{E}_{{z} \sim p_z({z})}[\log \big(1 - D(G({z} | {y}))\big)]
\end{aligned}
\end{equation}

$\mathcal{L}_{L_{2}}(G)$ is the pixel-wise MSE between actual MRI and the generated MRI. It is computed as:

\begin{equation} \label{l2G_equation}
\begin{aligned}
     \mathcal{L}_{L_{2}}(G) = \mathbb{E}_{x,y,z}[\norm{y-G(x,z)}_2]
\end{aligned}
\end{equation}

We have used also a modified variant of SSIM as a perceptual loss function. This function comes from Brunet \etal in~\cite{brunet_6059504} who studied the mathematical properties of SSIM. They develop a modified SSIM obeys that triangle inequality and so can be used as a distance function. They also suggest that a simpler version of this function, given in Eq.~\ref{ssim_loss_equation} below, very closely follows the provided distance function. This was the function we used as a perceptual loss function.
\begin{equation}  \label{ssim_loss_equation}
\begin{aligned}
     \mathcal{L}_{per}(G) = \mathcal{L}_{SSIM}(G) = \sqrt{1-SSIM(x,y)}
\end{aligned}
\end{equation}
Putting these components together, our complete loss function for the Generator is:
\begin{equation}  \label{loss_G_equation}
\begin{aligned}
     \mathcal{L}(G) = \arg\min_G\max_D \mathcal{L}_{cGAN}(G,D) + \\
                        \lambda \Big( \tau  \mathcal{L}_{L_{2}}(G) + (1 - \tau) \mathcal{L}_{per}(G)\Big)
\end{aligned}
\end{equation}
We use $\tau$ as a hyper-parameter to control the contribution of $\mathcal{L}_{L_{2}}(G)$ and $\mathcal{L}_{per}(G)$ in our total generator loss. $\mathcal{L}(G)$ defined in Eq.~\ref{loss_G_equation} is the our objective function for the generator to minimize. We have set $\lambda = 100$ in our experiments. We provide noise to the dropout layers during training of our MRI-GAN, instead of giving noise $z$ as an input to the generator as was experimented in~\cite{pix2pix2017}.

The objective function of the Discriminator model is given by:
\begin{equation}  \label{loss_D_equation}
\begin{aligned}
     \mathcal{L}(D) = \eta (\mathbb{E}_{f}[||f-\hat{f}||_2] + \mathbb{E}_{r}[||r-\hat{r}||_2])
\end{aligned}
\end{equation}
where, $f$ is the ground truth of fake samples, $\hat{f}$ is prediction on the fake samples by the discriminator, $r$ is the ground truth of real samples, $\hat{r}$ is prediction on real samples by the discriminator.
We set $\eta = 0.5$ to let discriminator learn more slowly than the generator. 

\subsection{MRI-GAN Architecture and Training}

Our generator model is based on a $n = 16$ layer, ``U-net'' architecture~\cite{ronneberger2015unet} with skip connections between each layer $i$ and layer $n-i$. We have $8$ down-sampling modules and $8$ up-sampling modules, in our generator model. Each down-sampling module is a sequence of Convolution2D-InstanceNorm2D-LeakyReLu-Dropout layers. Each up-sampling module is sequence of ConvTranspose2d-InstanceNorm2D-ReLU, except the last layer. The last layer module is of the form Upsample-ZeroPad2D-Convolution2D-Tanh~\cite{ioffe2015batch, xu2015empirical, 10.5555/3104322.3104425}.  Our discriminator model is series of Convolution2D-InstanceNorm2D-LeakyReLu layers. This discriminator follows the PatchGAN design described in~\cite{pix2pix2017}, and was chosen as it only penalizes structures such as the scale of image patches.  We started our MRI-GAN code development based off of the initial code from~\cite{github_PyTorch_GAN}. We used an Adam optimizer~\cite{kingma2017adam}, with a learning rate of $0.0002$, and chose momentum parameters $\beta_1=0.5$, $\beta_2=0.999$ for both the generator and the discriminator models. We trained our MRI-GAN with batch sizes of $128$. We had approximately $919,590$ DeepFake training and $1,359,717$ real training samples. We had approximately $339,930$ DeepFake test and $229,898$ real test samples in our MRI-DF dataset. Since we had fewer real samples as compared to DeepFake samples, for each epoch, we selected all fake samples and we randomly select an equal number of real sample from the dataset and we report results for each global batch of execution.

\subsection{Evaluation of MRI-GAN Results}

\begin{figure}[h!]
    \centering
    \includegraphics[width=0.50\textwidth,keepaspectratio]{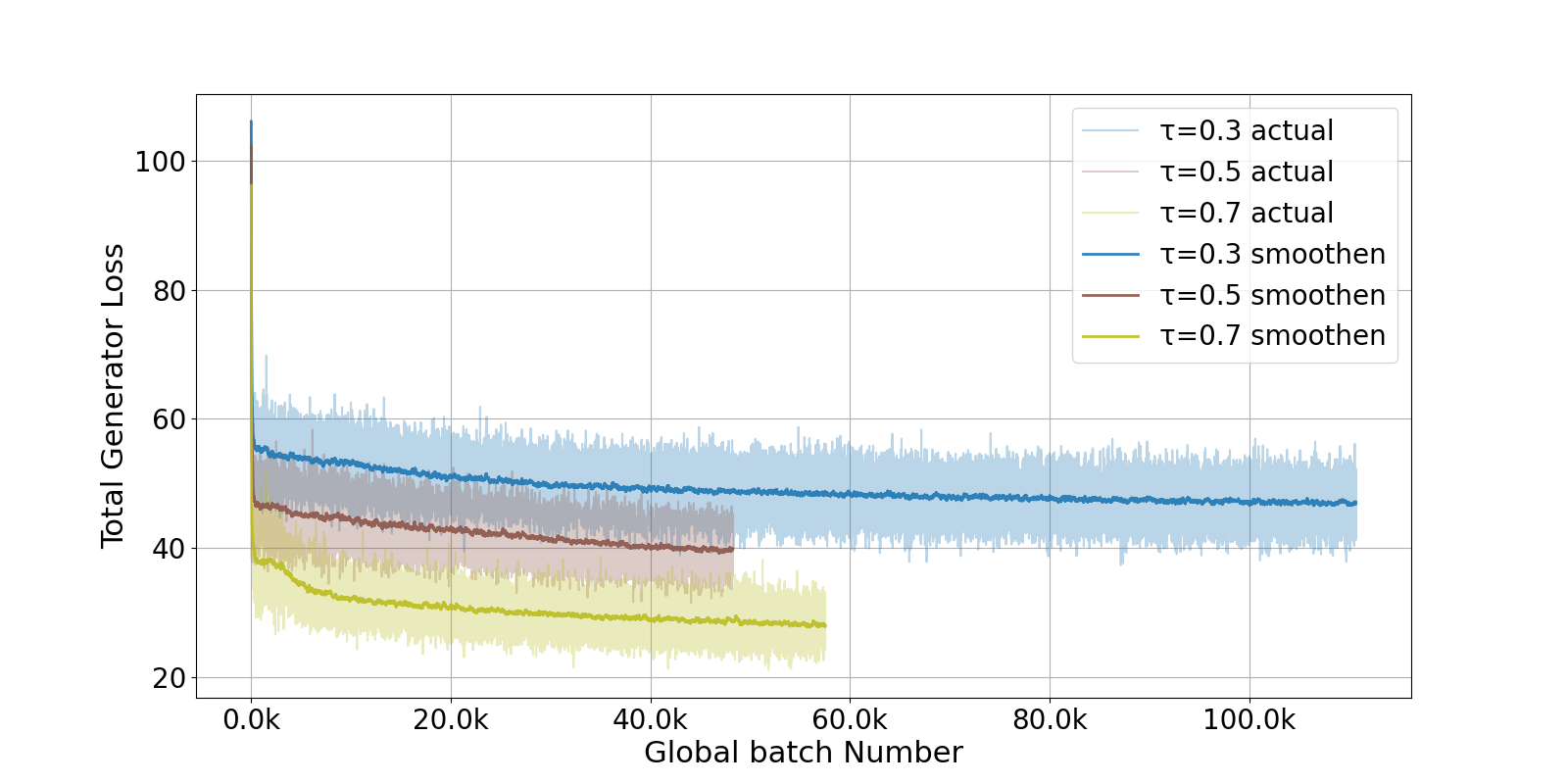}
    \caption{Total $\mathcal{L}(G)$ as a function of global batch number during training}
    \label{fig:loss_G_plot}
\end{figure}

\begin{figure}[h!]
    \centering
    \includegraphics[width=0.50\textwidth,keepaspectratio]{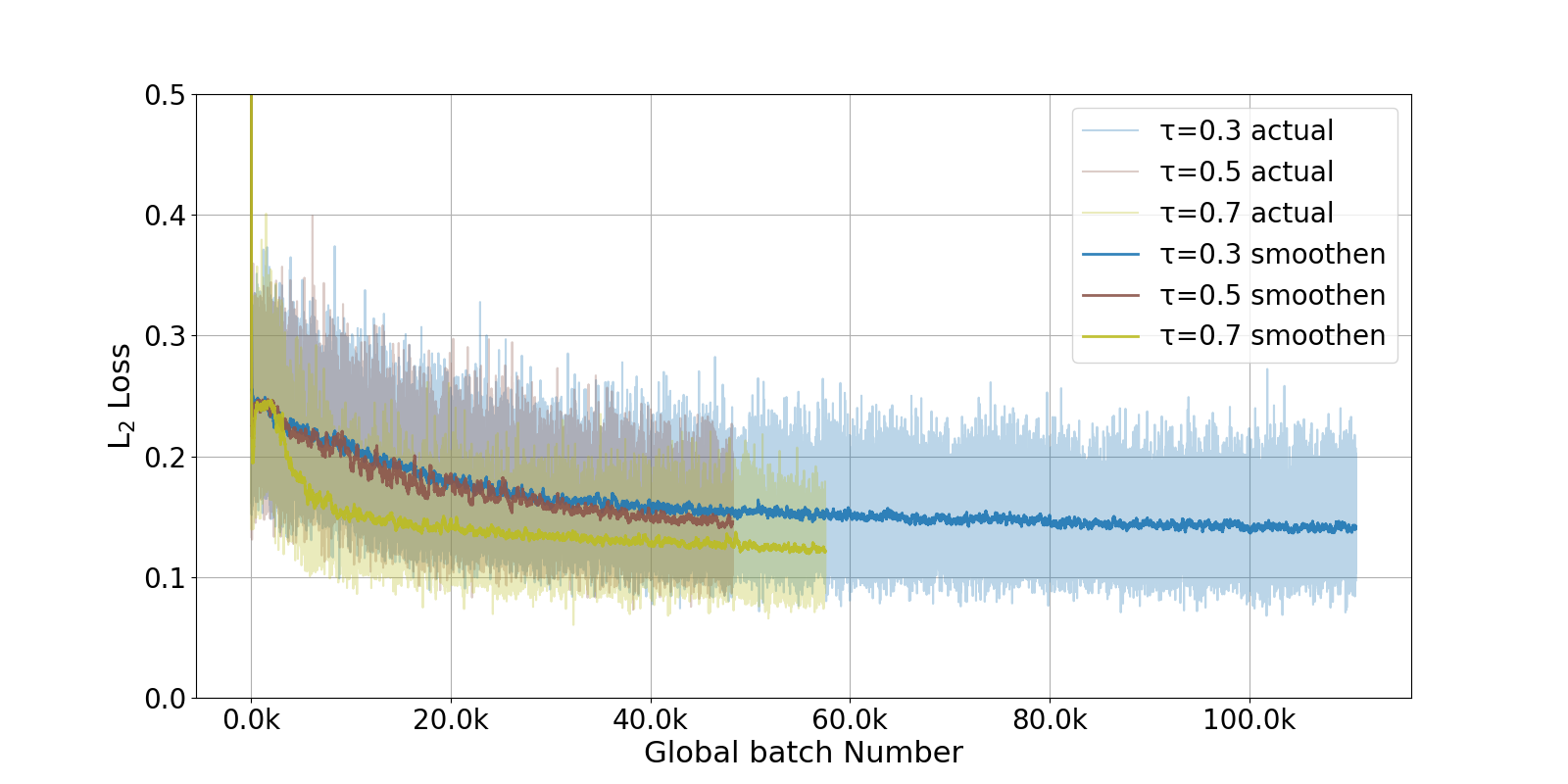}
    \caption{$\mathcal{L}_{L_{2}}(G)$ as function of global batch number during training}
    \label{fig:loss_l2_plot}
\end{figure}

\begin{figure}[h!]
    \centering
    \includegraphics[width=0.50\textwidth,keepaspectratio]{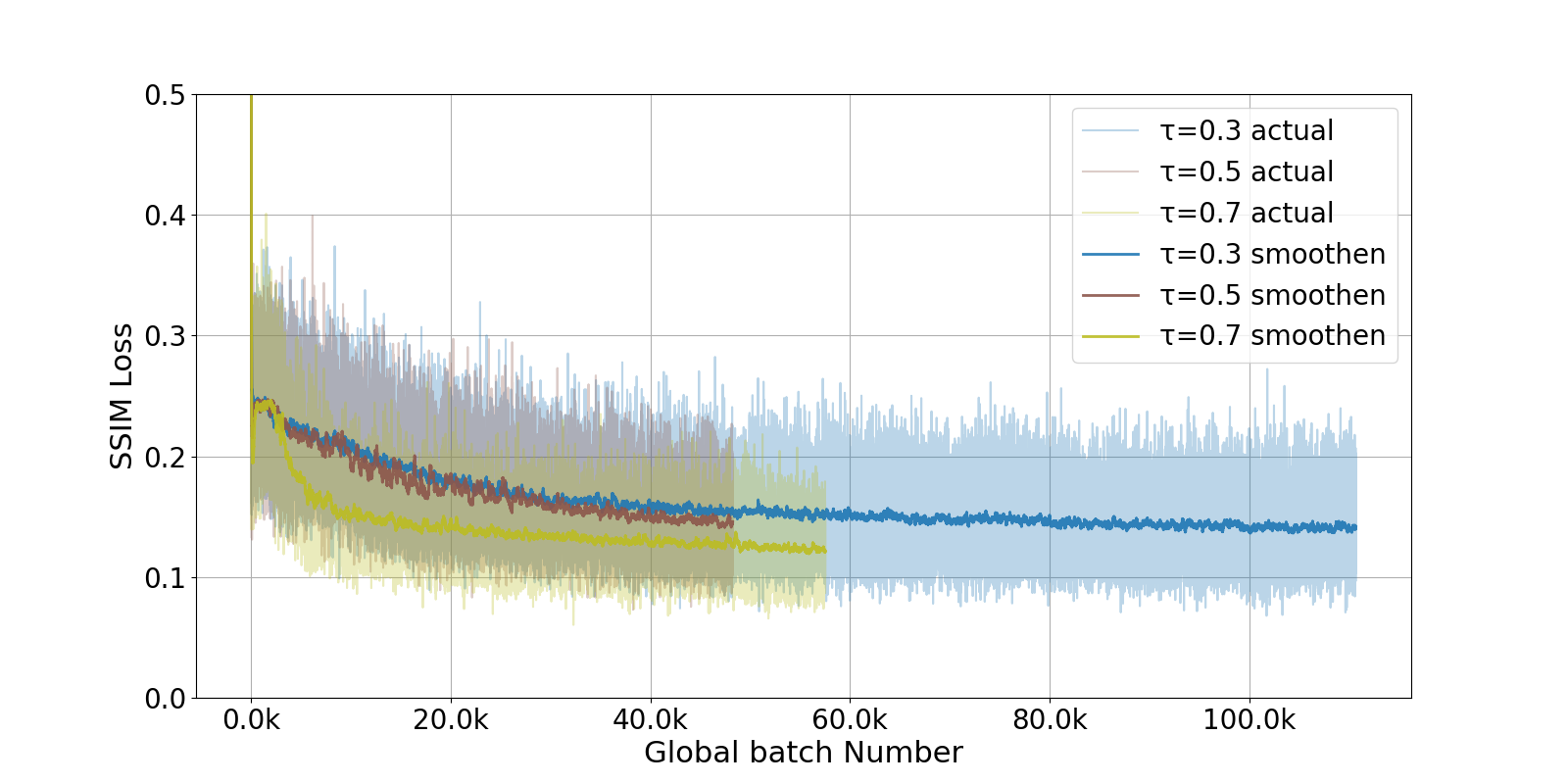}
    \caption{$\mathcal{L}_{SSIM}(G)$ as a function of global batch number during training}
    \label{fig:ssim_loss_plot}
\end{figure}

\begin{figure}[h!]
    \centering
    \includegraphics[width=0.50\textwidth,keepaspectratio]{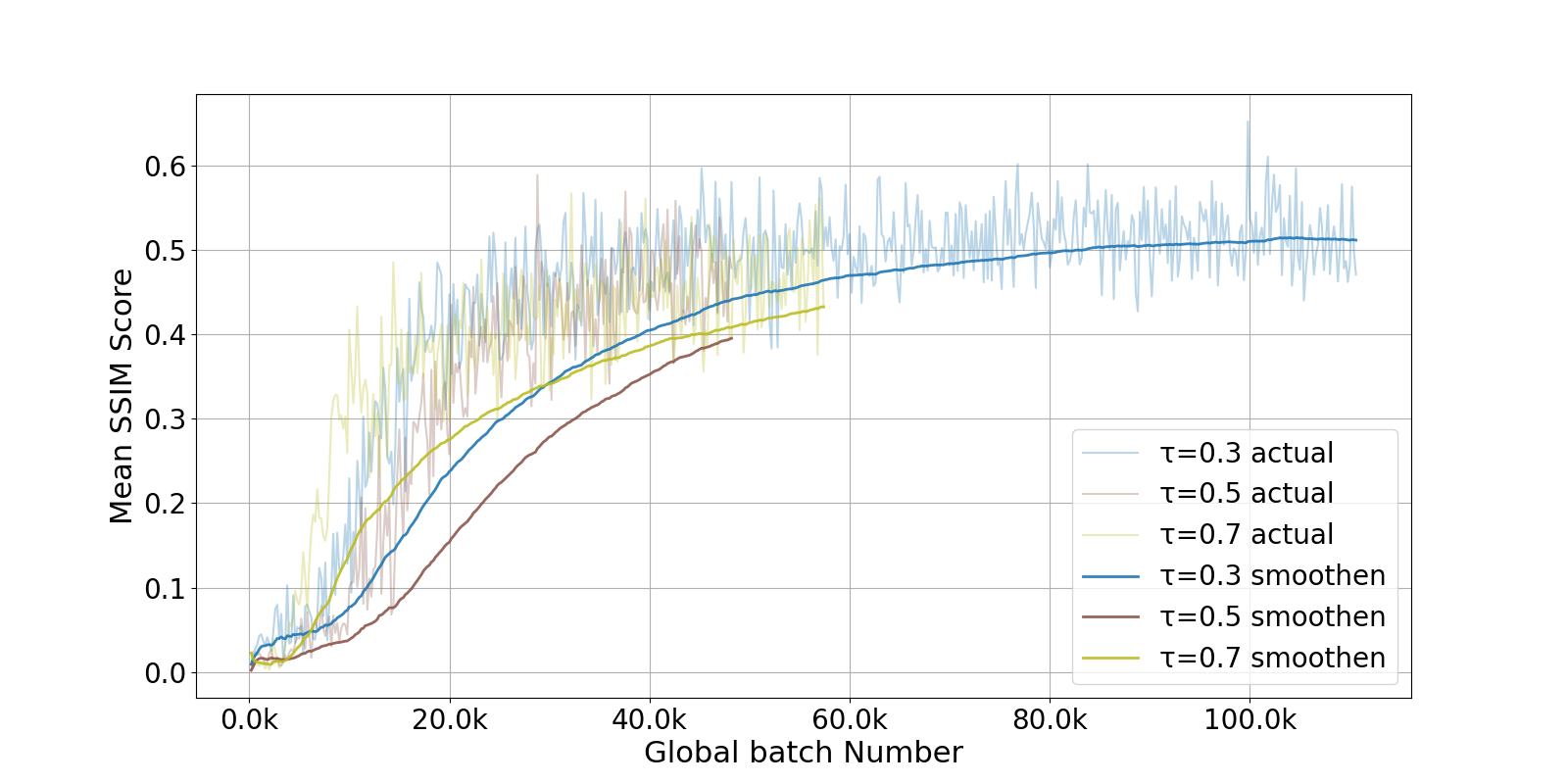}
    \caption{Mean SSIM score for samples images from test set of MRI-DF dataset as a function of global batch number during training.}
    \label{fig:ssim_score_plot}
\end{figure}

\begin{figure}[h!]
    \centering
    \includegraphics[width=0.50\textwidth,keepaspectratio]{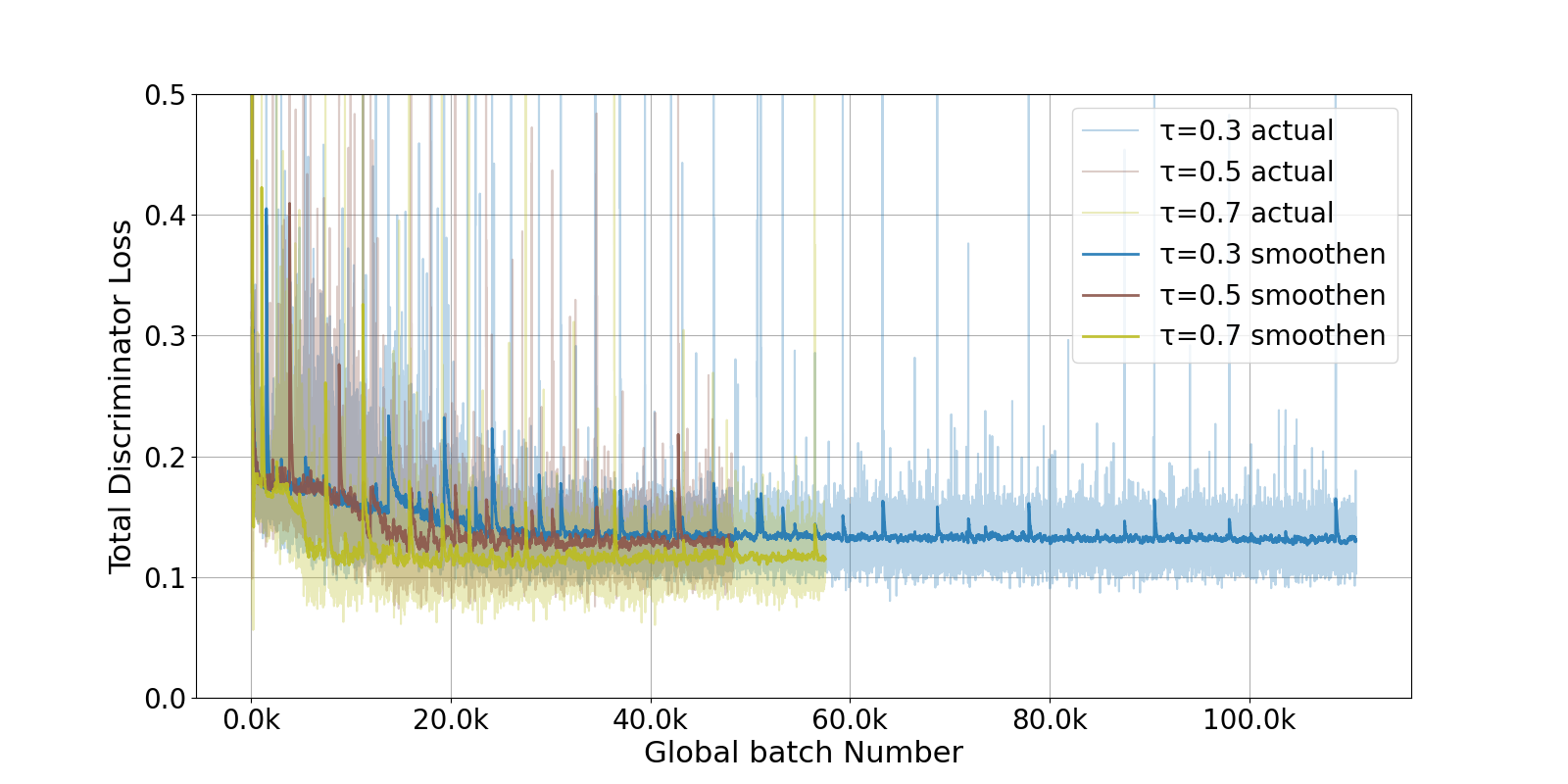}
    \caption{Total $\mathcal{L}(D)$ as a function of global batch number during training}
    \label{fig:loss_D_plot}
\end{figure}

\begin{figure*}[h]
    \centering
    \includegraphics[width=0.85\textwidth,height=\textheight,keepaspectratio]{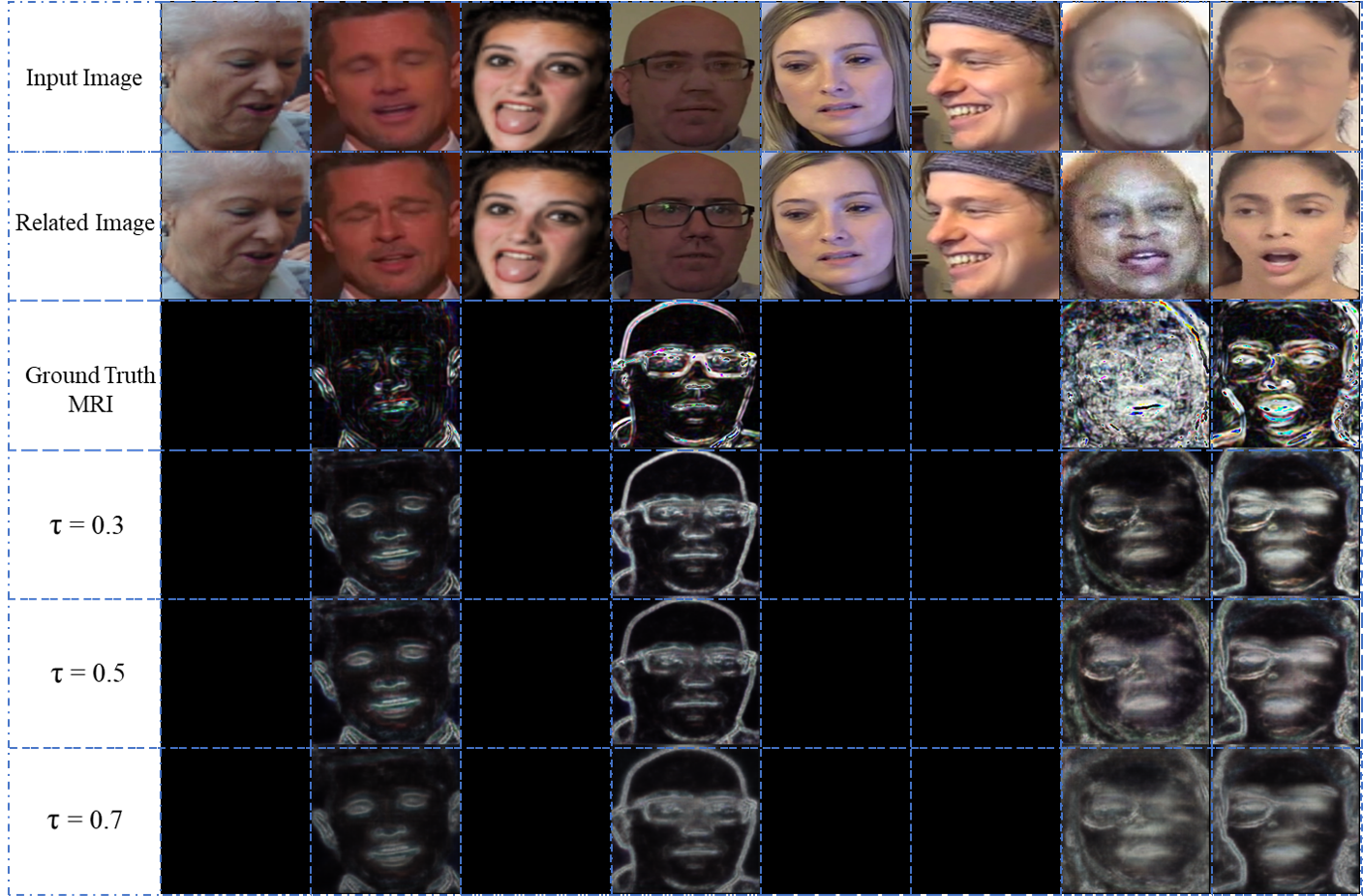}
    \caption{Sample MRIs generated by MRI-GAN with different values of $\tau$. The first row of images are input human faces, which can be either real or fake. The second row of images will be the same as the input image if it is not fake, otherwise, it is the real image the fake was based off of. Images in the third row are the the actual SSIM differences between the first two rows. It can be seen that if the face is synthesized, the pixel-wise SSIM is non-blank. The fourth, fifth, and sixth rows are the MRIs generated using MRI-GAN with $\tau$ of 0.3, 0.5, and 0.7 respectively.}
    \label{fig:MRI_demo_image}
\end{figure*}

Fig.~\ref{fig:loss_G_plot} shows the overall loss, given by Eq.~\ref{loss_G_equation}, for the generator model portion of our MRI-GAN. Fig.~\ref{fig:loss_l2_plot} and Fig.~\ref{fig:ssim_loss_plot} shows the L2 and SSIM loss, given by Eq.~\ref{l2G_equation} and Eq.~\ref{ssim_loss_equation}, respectively, for the generator model. Fig.~\ref{fig:loss_D_plot} shows the total loss of the discriminator. We did not train with $\tau = 0.7$ and $\tau = 0.5$ as long as we did for $\tau = 0.3$. During the training of our MRI-GAN, after every $200$th batch, we sampled $16$ images from the test set of the MRI-DF dataset. We generated the MRI images for this sample set using MRI-GAN and calculated the SSIM scores of these generated MRIs with respect to the ground-truths of MRIs. We took the mean value of all $16$ SSIM scores to quantify the learning progress of MRI-GAN. This mean SSIM scores are plotted in Fig.~\ref{fig:ssim_score_plot}. It can be seen that with $\tau = 0.3$ the SSIM score tends be better, so we chose this value for our final model. Fig.~\ref{fig:MRI_demo_image} shows the samples generated by MRI-GAN for various values of $\tau$.


\section{DeepFake detection}

Our DeepFake detection model takes $224 \times 224$ input  images. It then uses a pre-trained Efficient-Net B0 model to extract convolution features from these images. After this, adaptive average pooling is applied, the resulting tensors are flattened, and, finally, the resulting tensors are sent to a two layer, fully-connected, MLP. This MLP uses a dropout layer and a ReLU activation function between the two fully connected MLP layers. The final output of the model is a single neuron used to indicate if the input image is real or fake. For training, we used a binary cross-entropy loss function, we used an Adam optimizer, and we used a learning rate of 0.0001. For our dropout layers, we followed Srivastava \etal in ~\cite{10.5555/2627435.2670313}, and used $p = 0.5$. For regularization, we followed M\"{u}ller \etal in ~\cite{mller2019does} and used label smoothing with $\alpha = 0.1$.

As discussed in Section~\ref{section_methods}, we extract every tenth frame of each video. From each such frame we then extract the faces. All detected faces are passed to the DeepFake detection model to compute the probability of the face being fake.  Videos are made up of multiple frames and each frame may have zero or more faces. We use a simple aggregation algorithm to translate individual face-level classification data to entire video-level classification. We iterate over all faces detected in the video and count the number of faces whose probability of being fake is more than a constant FAKE\_FRAME\_THRESHOLD. If the number of such faces is more than a FAKE\_FRACTION of the total faces, we declare the video as fake, otherwise, we say it is real. We performed a grid-search to determine the best values of the FAKE\_FRAME\_THRESHOLD and FAKE\_FRACTION parameters.

\subsection{Results}

\begin{figure}[h!]
    \centering
    \includegraphics[width=0.45\textwidth,keepaspectratio]{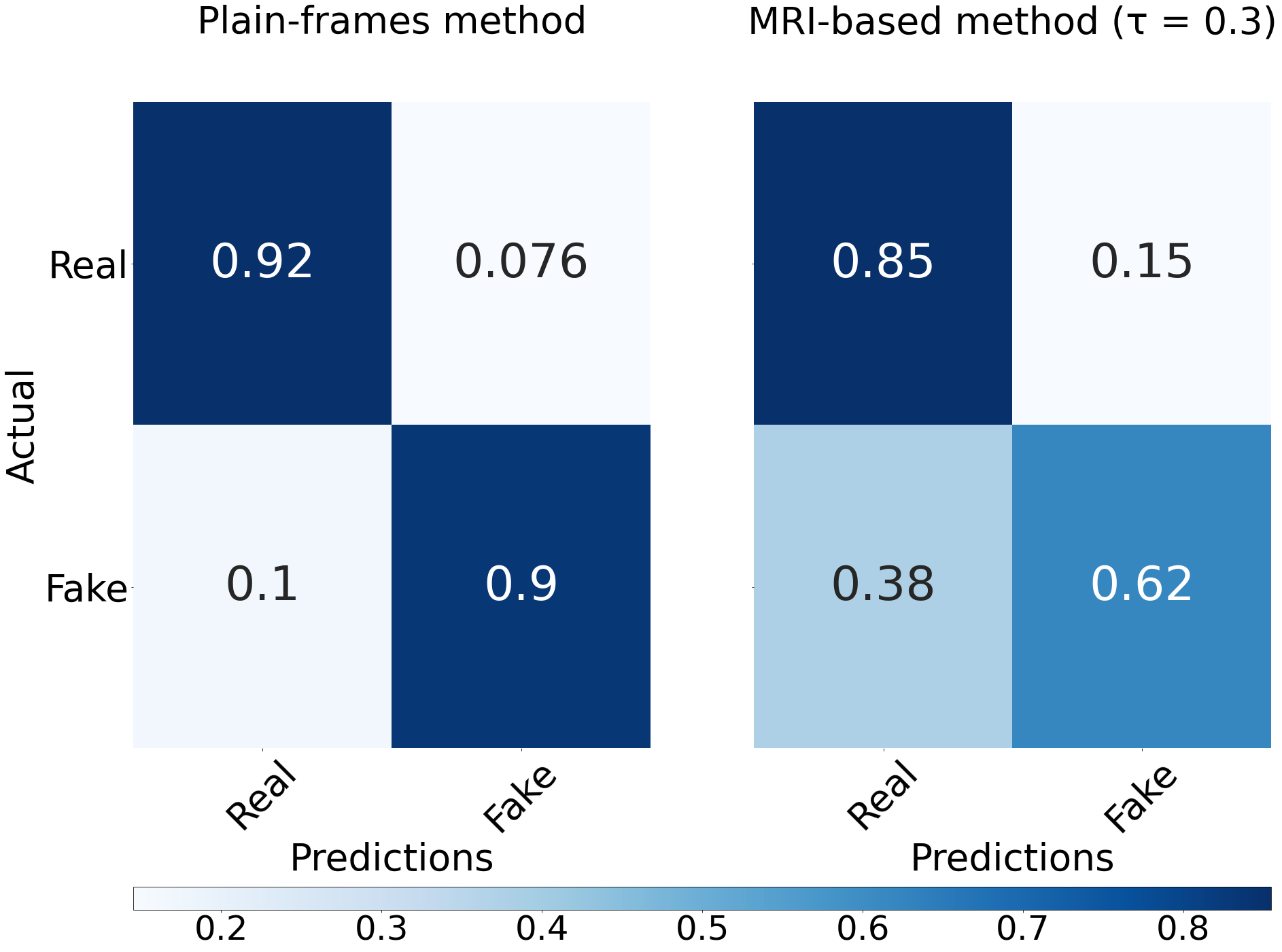}
    \caption{Confusion matrices of DeepFake detection.}
    \label{fig:conf_mat}
\end{figure}

Fig.~\ref{fig:conf_mat} shows the confusion matrices for our plain-frames and MRI-based detection methods. Table~\ref{tab:results-table} summarizes standard metrics used to quantify these techniques ability to detect DeepFakes. Table~\ref{tab:grid_param_results} shows the parameters we found using grid-search to translate face-level classification to video-level classification for both of these models. For the current set of experiments and tuning, our MRI-based method has a test accuracy of 74\% and plain-frames-based method achieves 91\% test accuracy.

\begin{table}[]
\label{tab:results-table}
\begin{tabular}{@{}lll@{}}
\midrule\midrule
 Metric   & \multicolumn{1}{c}{\begin{tabular}[c]{@{}c@{}}Plain-frame \\ method\end{tabular}} & \multicolumn{1}{c}{\begin{tabular}[c]{@{}c@{}}MRI-based \\ method \end{tabular}} \\ \midrule\midrule
True Positive Rate (TPR)           & 0.90                                                                               & 0.69                                                                             \\ \hline
False Negative Rate (FNR)          & 0.10                                                                               & 0.31                                                                             \\ \hline
False Positive Rate (FPR)          & 0.08                                                                               & 0.19                                                                             \\ \hline
True Negative Rate (TNR)           & 0.92                                                                               & 0.81                                                                             \\ \hline
Accuracy                           & 0.91                                                                               & 0.74                                                                             \\ \hline
Balanced Accuracy                  & 0.91                                                                               & 0.75                                                                             \\ \hline
F1-score                           & 0.91                                                                               & 0.77                                                                             \\ \hline
Precision                          & 0.92                                                                               & 0.85                                                                             \\ \hline
Specificity                        & 0.92                                                                               & 0.81                                                                             \\ \hline
Area Under the curve ROC           & 0.95	                                                                              & 0.76                                                                             \\
\midrule\midrule
\end{tabular}
\caption{Quantification of DeepFake detection results. MRI-based method is with $\tau$ = 0.3}
\end{table}

\begin{table}[]

\label{tab:grid_param_results}
\begin{tabular}{@{}lll@{}} 
\midrule\midrule
    \begin{tabular}[c]{@{}l@{}}DeepFake detection\\ method\end{tabular} & \multicolumn{1}{c}{\begin{tabular}[c]{@{}c@{}}FAKE\_FRAME\_\\ THRESHOLD\end{tabular}} & \multicolumn{1}{c}{\begin{tabular}[c]{@{}c@{}}FAKE\_\\ FRACTION\end{tabular}} \\ \midrule\midrule

plain-frames-based           & 0.80                                                                               & 0.30                                                                             \\ \hline
MRI-based ($\tau$ = 0.3)    & 0.70                                                                               & 0.30                                                                             
                                                                            \\
\midrule\midrule
\end{tabular}
\caption{Grid search parameters to translate face-level to video-level classification}
\end{table}

\section{Conclusion}

We propose a novel, GAN-based method for computing a likely pixel-wise perceptual dissimilarity of an input image from its related real image. We call this prediction an image MRI. We have developed a neural network to detect DeepFakes using these MRIs. In our experiments, we trained our MRI-GAN using SSIM as the perceptual dissimilarity function. Our approach is general, though, and could use other perceptual dissimilarity functions. Further, the method can be used to detect other kinds video manipulation, not just manipulations of faces. The DeepFake detection accuracy using MRI-GAN does not perform better than plain-frames-based methods that we also tested, however, we have some ideas and planned future work to improve the results of our MRI-based approach. These include trying other perceptual image assessment metrics such as those described in Zhai and Min~\cite{zhai2020perceptual} to train the MRI-GAN.  It also makes sense to try using an L1 loss function instead of an L2 loss function, as ~\cite{pix2pix2017} suggests L1 loss functions encourage less blurring effects in image-to-image translation.


{\small
\bibliographystyle{ieee_fullname}
\bibliography{main}
}

\end{document}